%% file: paper_draft.tex
\documentclass[11pt]{article}

\usepackage{amsmath,amssymb,amsthm}
\usepackage{graphicx}
\usepackage{hyperref}
\usepackage{natbib}
\usepackage[margin=1in]{geometry}
\usepackage{booktabs}
\usepackage{algorithm}
\usepackage{algorithmic}
\usepackage{xcolor}
\usepackage{subcaption}
\usepackage{placeins}

\title{Scale-Agnostic Kolmogorov-Arnold Geometry in Neural Networks}
\author{
    Mathew Vanherreweghe$^{1}$ \qquad
    Michael H. Freedman$^{2,3}$\thanks{Mathematical formulation and theoretical insights.} \qquad
    Keith M. Adams$^{1}$\thanks{Machine learning expertise and practical guidance.}
    \\
    \\
    $^{1}$Pebblebed \\
    $^{2}$Logical Intelligence \\
    $^{3}$Center of Mathematical Sciences and Applications, Harvard University \\
    \\
    \texttt{mathew@pebblebed.com}
}
\date{}

\begin{document}

\maketitle

\begin{abstract}
Recent work by Freedman and Mulligan demonstrated that shallow multilayer perceptrons spontaneously develop Kolmogorov-Arnold geometric (KAG) structure during training on synthetic three-dimensional tasks. However, it remained unclear whether this phenomenon persists in realistic high-dimensional settings and what spatial properties this geometry exhibits.

We extend KAG analysis to MNIST digit classification (784 dimensions) using 2-layer MLPs with systematic spatial analysis at multiple scales. We find that KAG emerges during training and appears consistently across spatial scales, from local 7-pixel neighborhoods to the full 28×28 image. This scale-agnostic property holds across different training procedures: both standard training and training with spatial augmentation produce the same qualitative pattern. These findings reveal that neural networks spontaneously develop organized, scale-invariant geometric structure during learning on realistic high-dimensional data.
\end{abstract}

\section{Introduction}

\subsection{The Kolmogorov-Arnold Theorem and Neural Networks}

In 1957, Andrey Kolmogorov and Vladimir Arnold proved a remarkable theorem about the representation of multivariate continuous functions \citep{kolmogorov1957representation, arnold1957functions}. The Kolmogorov-Arnold (KA) theorem states that any continuous function of multiple variables can be exactly represented as a finite composition and superposition of continuous functions of a single variable. More precisely, any continuous function $f: [0,1]^n \to \mathbb{R}$ can be written as:
\begin{equation}
f(x_1, \ldots, x_n) = \sum_{q=0}^{2n} \Phi_q\left(\sum_{p=1}^{n} \phi_{q,p}(x_p)\right)
\end{equation}
where $\Phi_q$ and $\phi_{q,p}$ are continuous univariate functions. This theorem provided a negative answer to Hilbert's thirteenth problem and revealed that high-dimensional functions possess a surprisingly low-dimensional  representation. This representation has the remarkable property that the inner functions $\phi_{q,p}$ can be chosen universally (that is, independent of $f$) and possess an idiosyncratic ``texture'', dubbed Kolmogorov-Arnold geometry (KAG) in \citet{freedman2025spontaneous}, and quantified through observables discussed below.

The connection to neural networks is one-to-one in the KAN construction \citep{liu2024kan}, but is also apparent in shallow multilayer perceptrons (MLPs). A standard two-layer network with activation function $\sigma$ computes:
\begin{equation}
f(x) = \sum_{i=1}^{h} v_i \, \sigma(w_i^T x + b_i)
\end{equation}
where $x \in \mathbb{R}^d$ is the input, $h$ is the hidden layer width, $w_i \in \mathbb{R}^d$ are weight vectors, and $v_i \in \mathbb{R}$ are output weights. The inner map $\phi(x) = \sigma(Wx + b) \in \mathbb{R}^h$ transforms the input into a hidden representation, while the outer map linearly combines these features. This architecture is, indeed, merely a simplified KA representation, suggesting that neural networks might organically discover KAG-like structure during training.

\subsection{Spontaneous Emergence of KA Geometry}

Recent work by \citet{freedman2025spontaneous} made a surprising discovery: vanilla MLPs trained on simple classification tasks spontaneously develop geometric signatures consistent with the Kolmogorov-Arnold representation. Analyzing the Jacobian matrix $J = \partial\phi/\partial x$ of the inner map, they identified three hallmark features of KAG:

\begin{enumerate}
    \item \textbf{Zero rows}: For a typical $x$, many hidden units become locally constant, with $\|\nabla \phi_i(x)\|_2 \approx 0$ where the row index $i$ depends on $x$.

    \item \textbf{Minor concentration} (MC): The distribution of $k \times k$ determinants (minors) of $J$ has a large spike at zero , but is simultaneously heavy-tailed, with a few remarkably large values. (MC is measured by three observables called: Participation ratio, KL-divergence, and RR.)
    \item \textbf{Alignment}: The Jacobian structure exhibits non-generic alignment that is not preserved under random orthogonal rotations
\end{enumerate}

These signatures emerge without explicit regularization or architectural constraints---they appear to be a natural consequence of gradient-based optimization. Freedman and Mulligan observed this phenomenon in shallow MLPs (single hidden layer, width $h=32$) trained on synthetic three-dimensional functions such as XOR, linear separators, and random Boolean functions.

This discovery suggests that KAG may represent a general organizational principle in neural networks, though it remained unclear whether these patterns persist beyond simple, low-dimensional settings.

\subsection{This Work}

While Freedman and Mulligan demonstrated KAG emerges in MLPs trained on synthetic 3D functions, their analysis left open questions about whether these patterns persist in high-dimensional, structured, real-world data. Moreover, their low-dimensional synthetic tasks did not permit investigation of spatial properties: questions about whether geometry varies across spatial regions or scales require data with richer spatial structure.

We extend their framework to MNIST digit classification (784-dimensional inputs, 10-class task) using 2-layer MLPs with varying capacity ($h \in \{64, 128, 256\}$). We confirm that KAG emerges in this realistic high-dimensional setting and introduce systematic spatial analysis to examine geometric structure at different scales.

We find that KAG appears consistently across spatial scales from local 7-pixel neighborhoods to the full 28×28 image. Most strikingly, this scale-agnostic property proves robust to training procedures: both standard training and training with spatial augmentation produce the same qualitative geometric patterns, though augmented training yields lower absolute participation ratios (~30\% reduction), consistent with reduced geometric 'frustration' from exposure to diverse data variations.

Section 2 describes our experimental methodology, Section 3 presents the main results, Section 4 discusses implications and connections to prior work, and Section 5 concludes.

\section{Methods}

We analyze KAG in 2-layer MLPs trained on MNIST digit classification through systematic spatial analysis at multiple scales. Complete implementation details, including all hyperparameters, hardware specifications, and reproducibility information, are provided in Appendix~\ref{appendix:experimental_details}.

We train 2-layer MLPs with GELU activations on MNIST digit classification ($28 \times 28$ grayscale images, 10 classes, 784-dimensional flattened input), varying the hidden dimension $h \in \{64, 128, 256\}$ to study how network capacity affects KAG. For both the first hidden layer map (L1) and second hidden layer map(L2), we compute Jacobians with respect to the input $x$ to examine how KA signatures evolve with training. Models are trained using AdamW for 200 epochs, with 5 random seeds per configuration for statistical robustness (error bars reflect $\pm 2$ standard deviations relative to these 5 runs). To test whether KAG is robust to training procedures, we train an additional set of models with the same configurations (5 seeds per hidden size) using spatial data augmentation with RandomAffine transformations (translation $\pm$4 pixels horizontal/vertical) applied during training.  Spatial analysis refers to the study of the scale at which KAG appears when analyzing minors based on pixels in both localized and well separated regions of the 2-dimensional input image.  This, of course, is quite distinct from the KAG of the first layer map from $\phi: \mathbb{R}^d \to \mathbb{R}^h$, and represents a new dimension of KAG exploration.

For each hidden layer representation $\phi: \mathbb{R}^d \to \mathbb{R}^h$, we compute the Jacobian matrix $J = \frac{\partial \phi}{\partial x} \in \mathbb{R}^{h \times d}$ using PyTorch's vectorized automatic differentiation. For the first hidden layer (L1), each of the $d=784$ columns corresponds to the derivative with respect to a specific input pixel in the flattened 28×28 image. For the second hidden layer (L2), we compute the Jacobian $\frac{\partial h_2}{\partial x}$ where $h_2$ represents the second layer activations. Through the chain rule, each of the $d=784$ columns captures how the second layer outputs change with respect to each input pixel, representing the end-to-end sensitivity of the network's internal representations to the input.

In \citet{freedman2025spontaneous}, the term KAG is defined in reference to the Jacobian of the first layer map. It is not a sharp, black-or-white criterion but rather a statistical tendency of the $\text{Jac}(x)$ to assume a form similar to the Jacobians of Kolmogorov and Arnold's first layer maps \citep{kolmogorov1957representation,arnold1957functions}. The idealized KAG of $\text{Jac}(x)$ is an $d \times h$ matrix in which each of the $h$ columns contains at most a single non-zero entry, where $d$ is the input dimension and $h$ is the number of hidden layer neurons. The observables detailed below (vanishing rows, PR, KL-divergence, and RR) quantify similarity to the KA ideal and contrariwise the departure from our untrained, random initializations.

To quantify the many-body character of KAG, we compute $k \times k$ determinants (minors) for $k \in \{1, 2, 3\}$ from the Jacobian. These minors measure the ``$k$-volume'' or determinant of a $k \times k$ submatrix of $\text{Jac}(x)$ formed by selecting some fixed set of $k$ columns and some fixed set of $k$ rows. We chose small k, because we wanted to avoid numerical issues with larger determinants, yet wanted to see a trend with increasing $k$. By design these minors are not rotationally invariant, but leverage the preferred neuron basis. We always use the absolute value of the minor as we have detected no signal in its sign, only its magnitude. Since exhaustively computing all minors is prohibitively expensive (e.g., $\binom{784}{3} \approx 79$ million 3-minors for MNIST), we randomly sample column combinations (up to 10,000 samples) and use adaptive chunking to manage GPU memory.

We quantify KAG using three metrics:
\begin{itemize}
    \item \textbf{Participation Ratio (PR)}: For a set of minor values $\{m_i\}$, the participation ratio is:
    \begin{equation}
    \text{PR} = \frac{\sqrt{\sum_i |m_i|^2}}{\sum_i |m_i|} = \frac{\|m\|_2}{\|m\|_1}
    \end{equation}
    A high PR indicates that few minors dominate (heavy-tailed distribution), consistent with KAG. We report the PR Ratio = $\text{PR}_{\text{trained}} / \text{PR}_{\text{random}}$, where values significantly greater than 1 indicate stronger concentration in trained networks.

    \item \textbf{KL Divergence}: We measure how much the minor distribution changes during training:
    \begin{equation}
    \text{KL}(P_{\text{trained}} \| P_{\text{init}}) = \sum_i P_{\text{trained}}(i) \log \frac{P_{\text{trained}}(i)}{P_{\text{init}}(i)}
    \end{equation}
    where $P$ denotes the normalized histogram of minor magnitudes. Large KL values indicate substantial distributional change. This measure was established as a reliable tool for detecting training induced shifts in minor concentration, correlating well with other metrics of KAG.(refer to \citet{freedman2025spontaneous} for details)

    \item \textbf{Rotation Ratio (RR)}: To test whether Jacobian structure is aligned with the neuron bases axis of $\mathbb{R}^d$ and $\mathbb{R}^h$, we apply random orthogonal transformations $J_{\text{rotated}} = J \cdot R$ where $R \in \text{SO}(h)$, computing:
    \begin{equation}
    \text{RR} = \frac{\mathbb{E}_R[\text{PR}(J \cdot R)]}{\text{PR}(J)}
    \end{equation}
    averaged over 100 random rotations. Values significantly greater than 1 indicate alignment (KA-like structure).
\end{itemize}

To test whether KAG appears at different spatial scales and locations, we perform three complementary analyses. First, for Euclidean ball sampling, we select varying radii $R \in \{7, 14, 21, 28\}$ pixels, randomly choose a seed pixel, and include all pixels within Euclidean distance $R$, producing approximately 150, 600, 1400, and 2500 pixels respectively. We then compute PR using only the Jacobian columns corresponding to these sampled pixels.

Second, for rectangular patch sampling, we partition the 28$\times$28 image into spatial regions (top-left, top-right, bottom-left, bottom-right, center) and compute PR separately for pixels in each region.

Third, to complement the localized euclidean ball analyses, we test whether geometry extends across space by constraining minor sampling to use only spatially well-separated pixels. When selecting $k$ columns (pixels) to form a $k \times k$ minor, we enforce that the selected pixels are pairwise separated by at least $D$ pixels (Euclidean distance), with $D \in \{0, 3, 5, 7, 10, 14\}$ pixels.

For all spatial analyses, we use 50 test images and report distributions across both images and random seed selections.

We compute all metrics at the end of training (epoch 200 for both standard and augmented models), comparing trained networks against random initialization baselines using the same architecture. We report mean $\pm$ 2 standard deviations across 5 random seeds for the $h \in \{64, 128, 256\}$ configurations.

\section{Results}
\label{sec:results}
Let us say concretely: We demonstrate that KAG emerges during training in 2-layer MLPs on MNIST, appearing consistently across spatial scales from local neighborhoods to the full image. This scale-agnostic property is robust to training procedures, appearing in both standard and augmented training regimes.

\begin{figure}[!htb]
\centering
\begin{subfigure}[b]{0.47\textwidth}
    \centering
    \includegraphics[width=\textwidth]{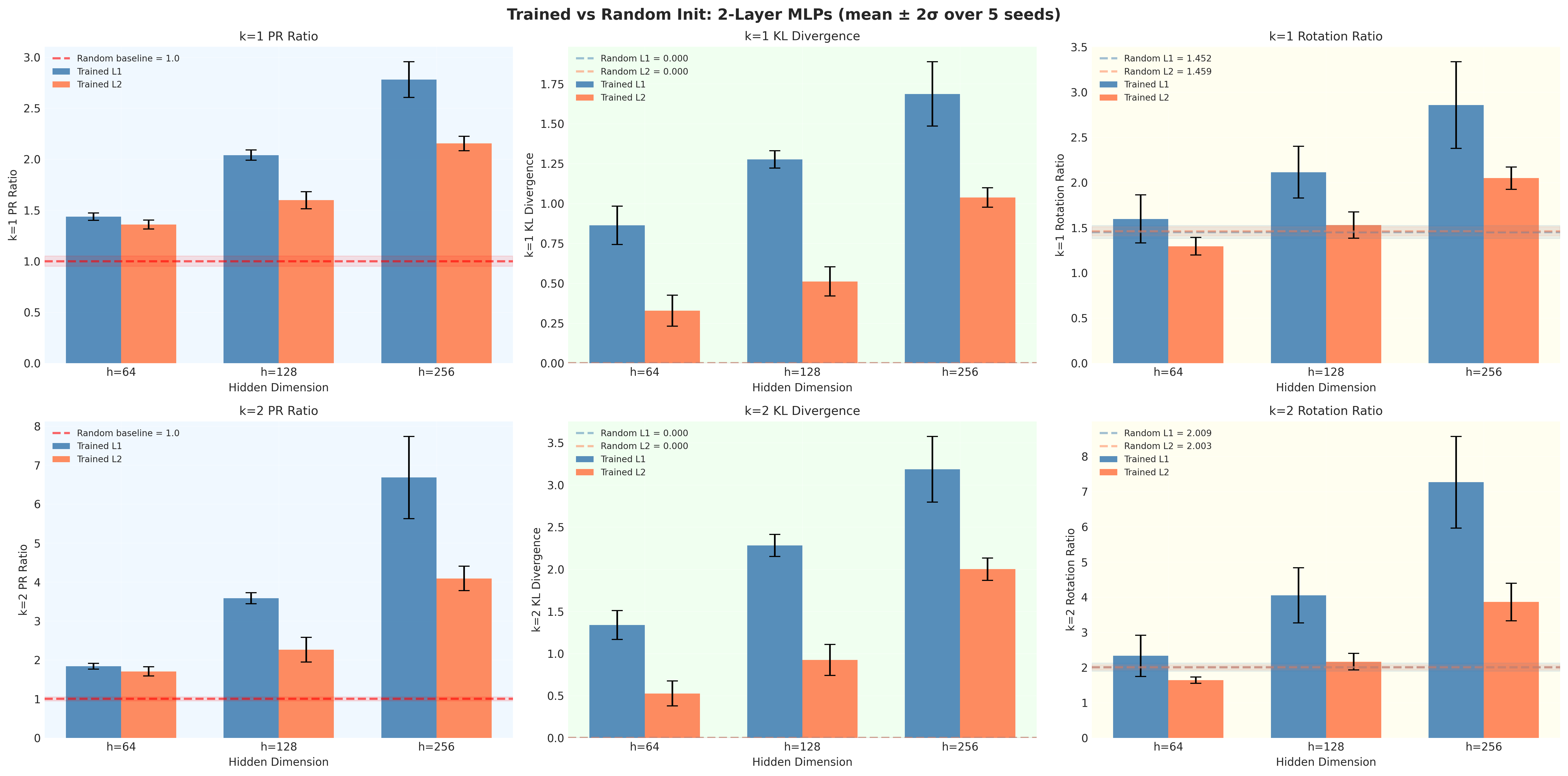}
    \caption{Standard model}
    \label{fig:trained_vs_random_standard}
\end{subfigure}
\hfill
\begin{subfigure}[b]{0.47\textwidth}
    \centering
    \includegraphics[width=\textwidth]{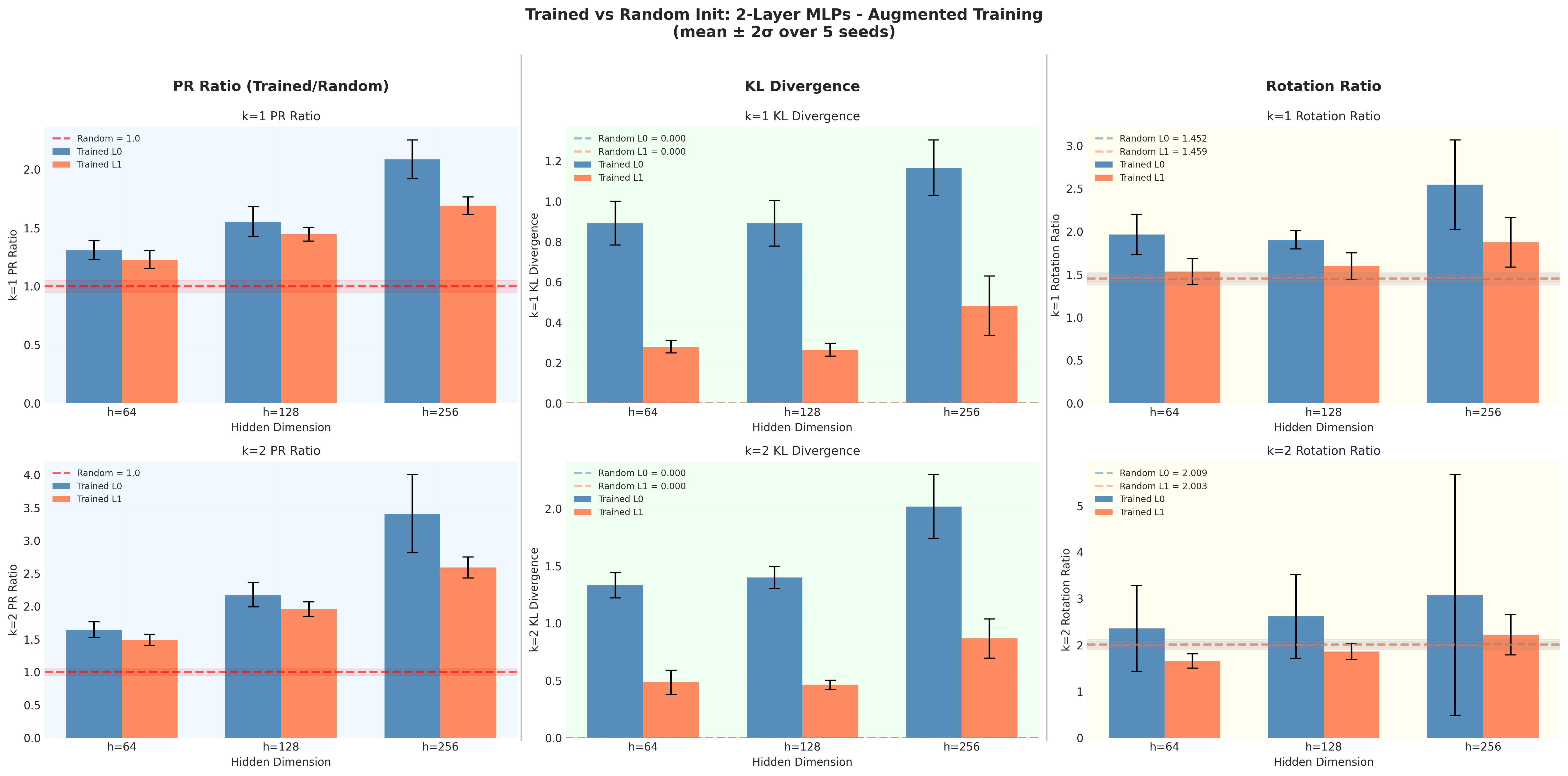}
    \caption{Augmented model}
    \label{fig:trained_vs_random_augmented}
\end{subfigure}
\caption{KAG emerges during training for both standard and augmented models. First Layer map (blue bars) and second layer map (orange bars) across three KAG metrics: PR ratio, KL divergence, and RR. Results shown for minor orders $k \in \{1,2\}$ with hidden dimensions $h \in \{64, 128, 256\}$. Red horizontal lines and shaded regions (mean $\pm$ 2 standard deviations) indicate baseline values for these metrics across 5 random seeds. Trained networks consistently exceed random baselines, with larger networks showing stronger geometric signatures. Both training procedures show the same qualitative patterns. Note that large RR error bars in the augmented model (k=2) are due to one seed with exceptionally high values, not bidirectional variance.}
\label{fig:trained_vs_random}
\end{figure}

Figure~\ref{fig:trained_vs_random} compares trained networks against random initialization across three metrics (PR, KL divergence, RR) for varying hidden dimensions ($h \in \{64, 128, 256\}$) and minor orders ($k \in \{1, 2\}$) for both standard and augmented training. Trained networks consistently show markedly stronger KA signatures than random initialization across all metrics, with larger networks exhibiting more pronounced geometric structure. Layer 1 shows stronger signatures than layer 2, and higher-order minors ($k=2$) reveal stronger effects than $k=1$.

The larger KAG signatures for L1 compared to L2 supports the hypothesis that KAG is primarily about \emph{data preparation}---the first layer arranges data propitiously for later analysis. This is consistent with KA signatures diminishing once the network has been fully traversed. Since L2 is a composition of a linear output map with L1, we expect to see some residual signatures, but a linear map can only do so much to preserve or enhance geometric structure. 

Notably, all model sizes achieved similar classification performance ($\sim$98\% test accuracy), indicating that the differences in KAG across network capacities are not driven by differences in task performance. MNIST is a sufficiently simple task that even the smallest network ($h=64$) learns the classification problem effectively, yet larger networks develop stronger geometric signatures during this learning process. To put this in perspective, even our largest hidden dimension ($h \leq 256$) are all smaller than the input dimension ($d=784$), opposite to the original KA construction where $h \geq 2d+1$.

\subsection{Spatial Properties of KA Geometry}

Having established that KAG emerges during training, we investigate at what spatial scales it appears.

Before presenting spatial analysis results, we clarify the three distinct spaces involved in our analysis. First, there is the input space (784-dimensional flattened MNIST images). Second, there is the hidden space where the network's learned representations live ($h$-dimensional, $h \in \{64, 128, 256\}$). We compute Jacobians for both hidden layers (L1 and L2) with respect to the input to examine how KA signatures evolve with depth. KAG concerns the local properties of these maps from input to hidden space, specifically the structure of the Jacobian matrices $\text{Jac}(h(x))$ at typical inputs $x$. Third, there is the spatial structure of the MNIST image itself (28×28 pixel grid). Our spatial analyses probe how KAG (which describes the input to hidden maps) varies when we sample Jacobian columns corresponding to different spatial regions of the MNIST image. This allows us to ask how KAG varies across different spatial regions of the image.

We now examine whether KAG appears uniformly across spatial scales or varies with the region sampled from the MNIST image. To ensure our findings are not artifacts of training procedure, we compare both standard and augmented models.

\begin{figure}[htbp]
\centering
\begin{subfigure}[b]{0.39\textwidth}
    \centering
    \includegraphics[width=\textwidth]{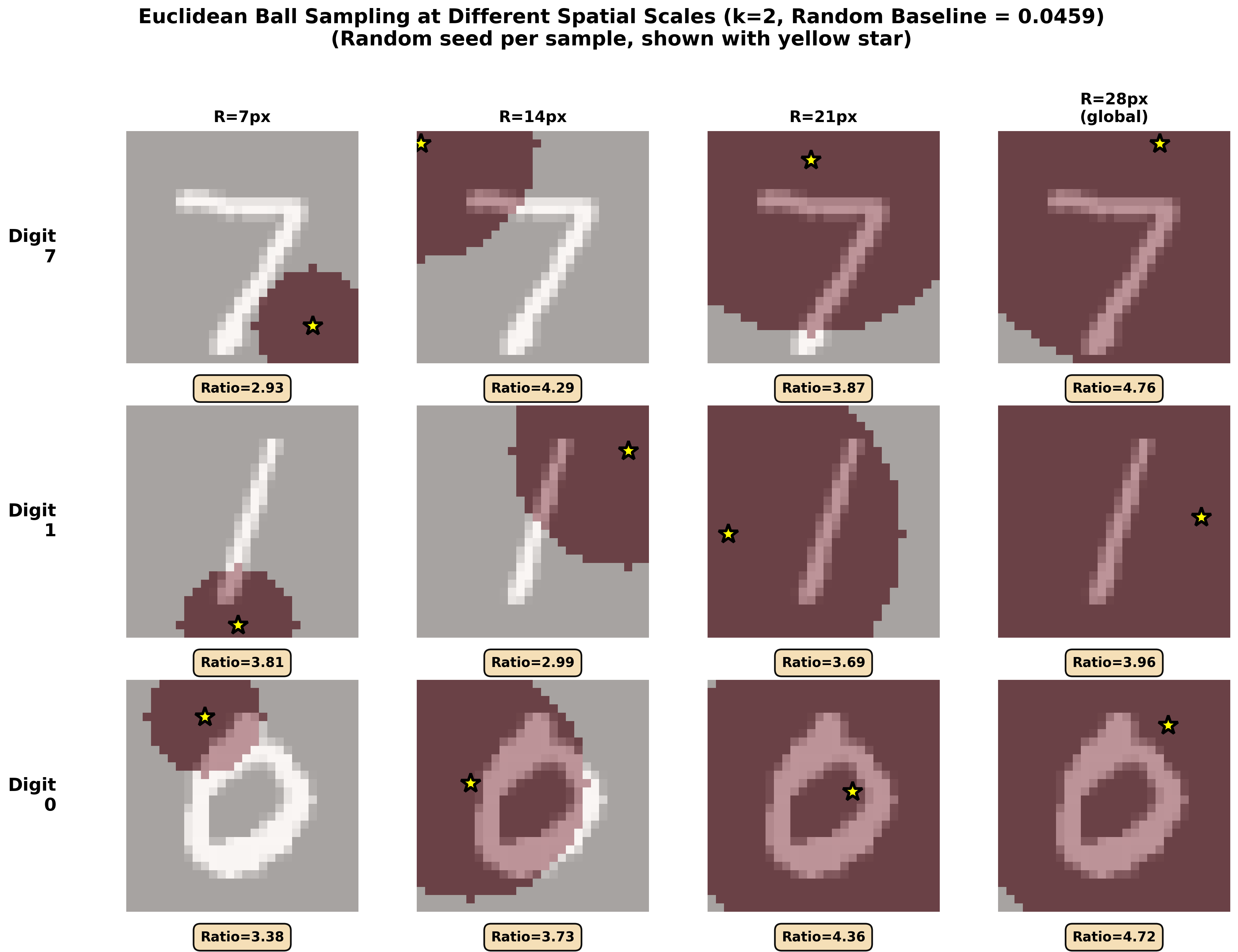}
    \caption{Euclidean ball sampling illustration}
    \label{fig:euclidean_illustration}
\end{subfigure}
\hfill
\begin{subfigure}[b]{0.39\textwidth}
    \centering
    \includegraphics[width=\textwidth]{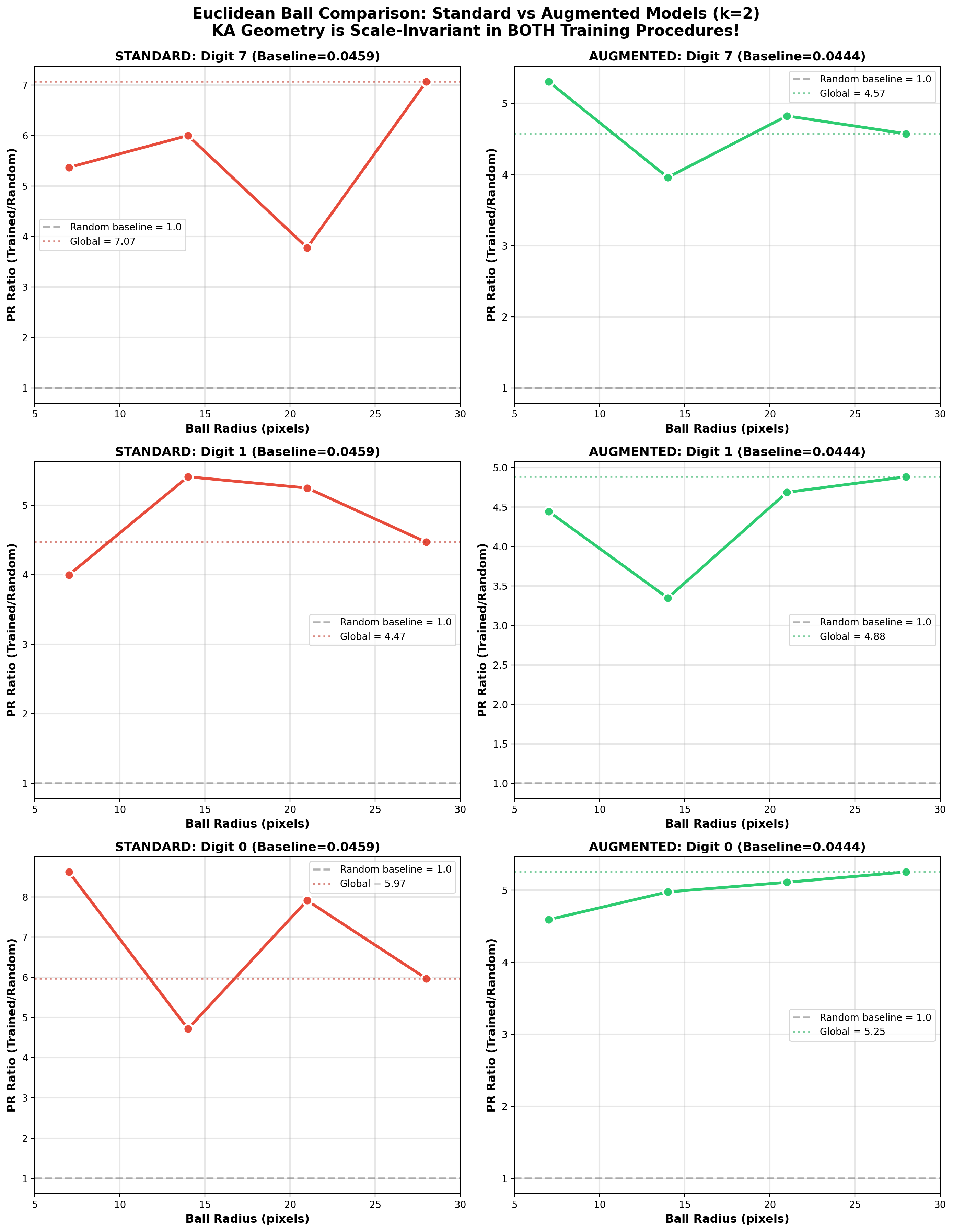}
    \caption{Scale-agnostic geometry comparison}
    \label{fig:euclidean_comparison}
\end{subfigure}
\caption{Scale-agnostic KAG across spatial scales. (a) Illustration of Euclidean ball sampling at radii $R \in \{7, 14, 21, 28\}$ pixels. Yellow stars mark random seed pixels, with sampled regions highlighted. (b) Both standard and augmented models exhibit scale-agnostic KAG: PR ratios remain well above baseline (1.0) across all ball radii for k=2 minors. Augmented model exhibits lower PR ratios (approximately 30\% reduction), consistent with reduced 'frustration' from exposure to diverse data variations.}
\label{fig:euclidean_analysis}
\end{figure}

First, we test local to medium scales by sampling minors from pixels within varying radii $R \in \{7, 14, 21, 28\}$ pixels from random seed points. Figure~\ref{fig:euclidean_illustration} illustrates this sampling procedure for different radii. Figure~\ref{fig:euclidean_comparison} shows that both models exhibit scale-agnostic KAG: PR ratios remain well above baseline (1.0) across all radii for both training procedures. Notably, the augmented model shows approximately 30\% lower PR ratios while maintaining the same scale-invariant pattern, consistent with reduced geometric 'frustration' from exposure to diverse data variations during training.

\begin{figure}[htbp]
\centering
\begin{subfigure}[b]{0.39\textwidth}
    \centering
    \includegraphics[width=\textwidth]{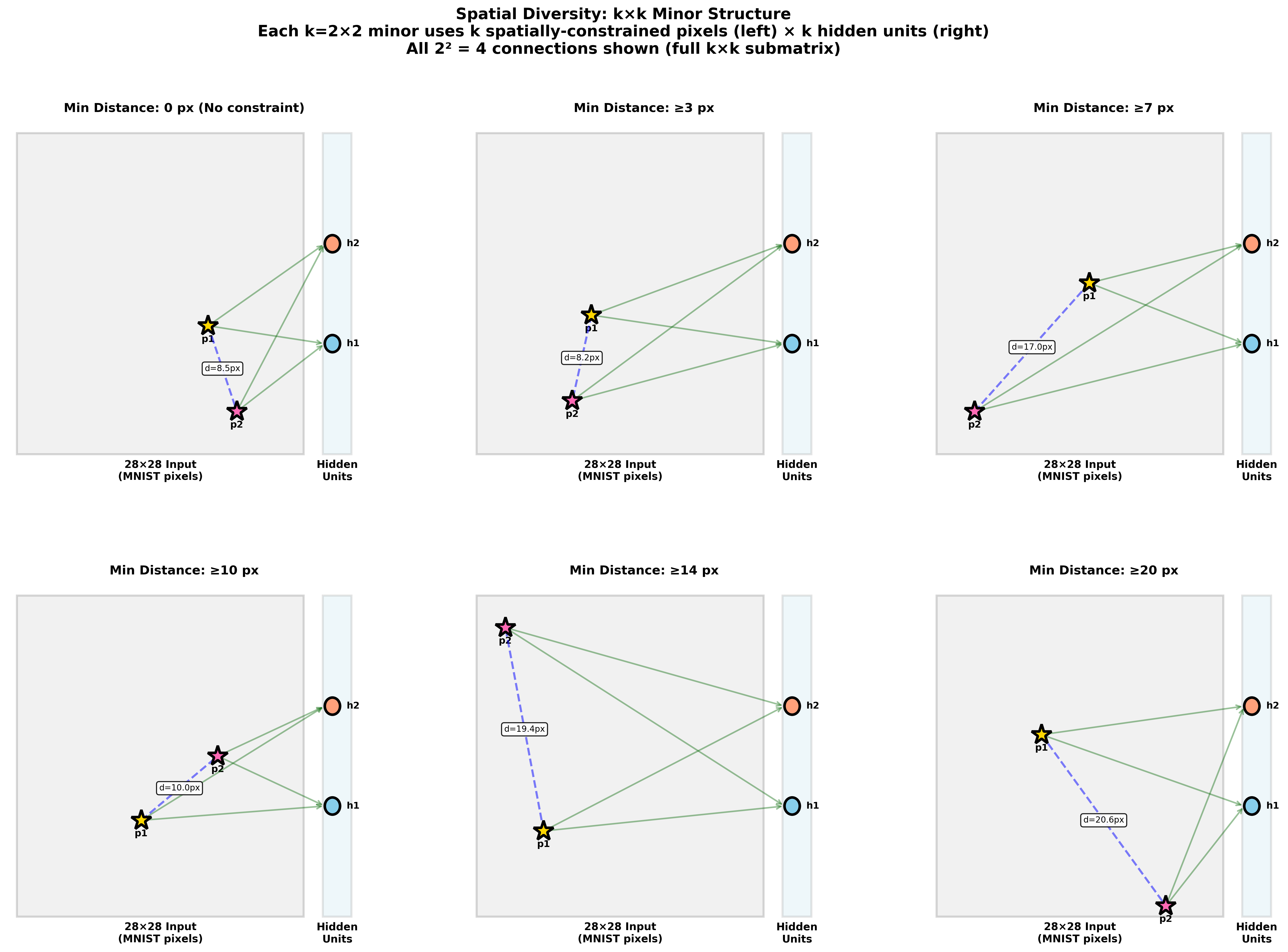}
    \caption{Distance constraint illustration}
    \label{fig:distance_illustration}
\end{subfigure}
\hfill
\begin{subfigure}[b]{0.39\textwidth}
    \centering
    \includegraphics[width=\textwidth]{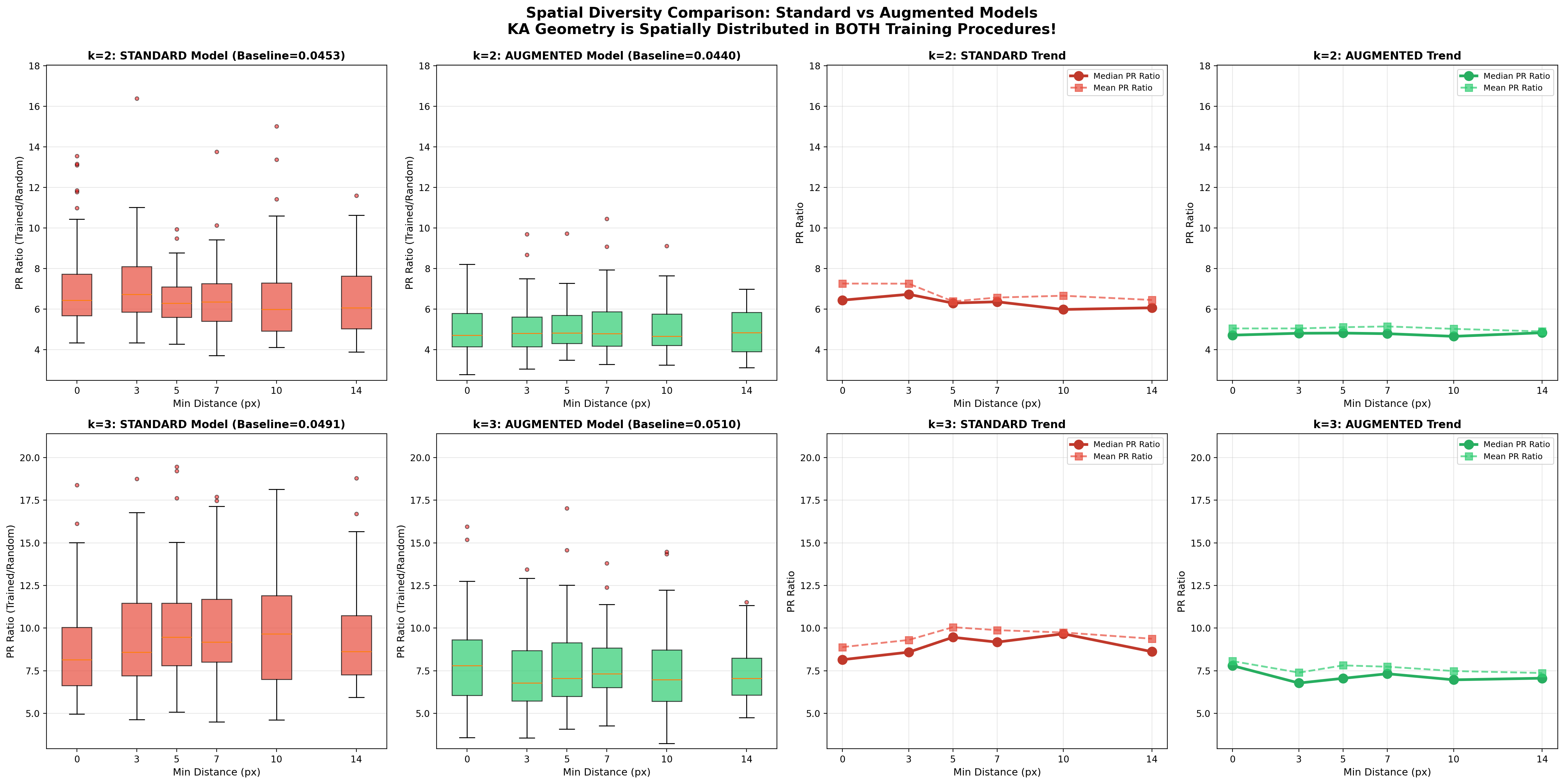}
    \caption{Global geometry comparison}
    \label{fig:spatial_diversity_comparison}
\end{subfigure}
\caption{Spatially distributed KAG across training procedures. (a) Illustration of minimum distance constraint for $k=2$ pixel pairs. Colored lines connect sampled pairs with distance labels. As minimum distance increases from 0px to 14px, fewer valid pairs exist, but those that do remain spatially distributed. (b) Both standard and augmented models maintain stable PR ratios across distance constraints. Box plots show PR ratio distributions for both k=2 (top) and k=3 (bottom) minors, with trend lines showing median and mean stability.}
\label{fig:spatial_diversity_analysis}
\end{figure}

Second, we test whether geometry extends globally by constraining minors to well-separated pixels with minimum distance $D \in \{0, 3, 5, 7, 10, 14\}$ pixels (Figure~\ref{fig:distance_illustration}). Figure~\ref{fig:spatial_diversity_comparison} shows PR ratios remain stable even at 14 pixels apart (half the image's linear dimension) for both models and both $k=2$ and $k=3$ minors, demonstrating KAG extends across the full input space regardless of training procedure.

\FloatBarrier

\section{Discussion}
\label{sec:discussion}

While our study is fundamentally observational, the patterns we observe connect to several strands of recent work on neural network structure. KAN networks \citep{liu2024kan} explicitly engineer Kolmogorov-Arnold representations using learnable univariate activations, treating KA structure as an architectural choice. The fact that vanilla MLPs spontaneously develop similar geometry during training suggests something different: perhaps KA structure is less about architectural design and more about what gradient descent naturally discovers when learning complex functions. This echoes the broader theme in deep learning of learned representations often matching or exceeding bespoke biases when sufficient data and capacity are available.

An intriguing observation emerges when comparing models trained with and without spatial augmentation. While both develop scale-agnostic KAG, the augmented model exhibits approximately 30-40\% lower participation ratios despite maintaining comparable classification accuracy (96\% vs 98\%). This aligns with the notion of geometric 'frustration' where models trained on limited data variations (centered digits only) develop more extreme geometric structure as they struggle to learn from constrained distributions. Augmented training exposes the network to diverse spatial configurations, apparently reducing the need for such extreme geometric organization. This connects to frustration experiments from \citet{freedman2025spontaneous} showing that networks trained on random or blank data develop extremely high PR values as they attempt to extract patterns from noise. The relationship between training distribution diversity and geometric structure warrants systematic investigation.

The relationship to neural collapse \citep{papyan2020prevalence} is particularly intriguing. Neural collapse describes how class-level representations become increasingly structured late in training, with class means collapsing toward symmetric simplex configurations. Our work operates at a different scale: we examine the Jacobian mapping from inputs to hidden representations rather than final-layer class structure. It's tempting to speculate that these phenomena might be connected. Where KAG in early layers provides the structured substrate that enables neural collapse in later layers, or conversely, that the pull toward collapsed final representations induces geometric organization earlier in the network. But these remain open questions requiring careful analysis of the joint evolution of geometry across all layers during training.

Of course, our analysis has certain limitations. We study only MNIST, a relatively simple grayscale recognition task, though this does represent a substantial step up in complexity from the 3D synthetic functions analyzed in prior work \citep{freedman2025spontaneous}. Our networks are shallow (just two hidden layers) and whether these patterns persist in modern deep architectures remains to be tested. Further, while the RR metric shows clear signal, it exhibits higher baseline variance than participation ratio due to max-sampling effects, making it better suited for secondary confirmation than primary claims. 

This points to clear next steps. The most critical is moving beyond correlation to causation. Frustration experiments (training on random labels before fine-tuning on real data) could test whether pre-established KAG accelerates learning. Explicit regularization experiments, adding loss terms that encourage or discourage KA signatures, could probe whether geometric structure actively aids generalization or merely accompanies it. These interventions would address the core question: Does KAG cause effective learning, or does learning cause KAG, or both?

Beyond causality, extending this analysis to modern architectures would clarify whether KAG is specific to fully-connected networks or a more universal principle. Comprehensive analysis across CNNs, ResNets, and attention-based models is needed to determine whether these geometric patterns are architecture-specific or general. Scaling to larger datasets (ImageNet for vision, language modeling datasets for transformers) would test whether these geometric patterns persist in contemporary settings or remain an artifact of simple tasks. Theoretical work explaining why gradient descent discovers KA structure could connect to broader questions about implicit regularization and the geometry of loss landscapes.

Ultimately, these results suggest that the geometric structure of learned representations may be richer and more systematic than previously appreciated. The spontaneous emergence of Kolmogorov-Arnold patterns, their consistency across spatial scales, and their correlation with network behavior all point toward geometric organization as a potentially important lens for understanding how neural networks learn. Whether this geometry is merely an interesting mathematical curiosity or plays a functional role in learning and generalization is an exciting next step.

\section{Conclusion}
\label{sec:conclusion}

We have extended the analysis of Kolmogorov-Arnold geometry from synthetic 3D functions to realistic 784-dimensional MNIST classification, revealing that this geometric structure emerges spontaneously during training and exhibits rich organizational properties. Unlike prior work on small synthetic datasets, our experiments on a high-dimensional classification task demonstrate that KAG is fundamentally scale-agnostic: it appears consistently from local neighborhoods as small as 7 pixels to the full 28×28 image, and extends globally across the input space even when constrained to well-separated pixels.

Perhaps most strikingly, this scale-agnostic property proves robust to training procedures. Both standard training and training with spatial augmentation produce the same qualitative geometric patterns, though augmented training yields lower PR ratio. This suggests that training distribution diversity modulates the strength of geometric structure---networks exposed to greater data variation require less extreme geometry to achieve comparable performance.

The spontaneous emergence of scale-invariant geometric structure is particularly intriguing. It suggests that KAG may reflect fundamental properties of how gradient descent organizes representations in overparameterized networks, rather than being specific to particular training regimes or datasets. Whether this structure actively aids learning or emerges as a consequence of optimization dynamics remains an important open question that intervention experiments could address.

Looking forward, extending this analysis to modern architectures and larger-scale tasks would clarify whether scale-agnostic KAG is a general principle of learned representations or specific to fully-connected networks on simple vision tasks. Systematic investigation of how training procedures shape geometric structure could yield insights into implicit regularization and inform the design of training protocols that encourage salubrious geometric properties.

\section*{Acknowledgements}

We thank Tammie Siew and Pamela Vagata (Pebblebed) for invaluable discussions that helped clarify the conceptual foundations of this work and sharpen our presentation of the key ideas.

\bibliographystyle{plainnat}
\bibliography{references}

\input{appendix}

\end{document}

%% file: appendix.tex
\appendix

\section{Detailed Experimental Configuration}
\label{appendix:experimental_details}

This appendix provides complete implementation details for all experiments reported in the main paper, enabling full reproducibility of our results.

\subsection{MNIST MLP Training Details}

\subsubsection{Architecture Specifications}

Our 2-layer MLPs follow the architecture:
\begin{equation}
f(x) = W_{\text{out}} \sigma(W_1 \sigma(W_0 x + b_0) + b_1)
\end{equation}
where $x \in \mathbb{R}^{784}$ is a flattened MNIST image, $W_0 \in \mathbb{R}^{h \times 784}$ and $W_1 \in \mathbb{R}^{h \times h}$ are weight matrices for the two hidden layers, $\sigma$ is the GELU activation function, and $W_{\text{out}} \in \mathbb{R}^{10 \times h}$ maps to 10 output classes.

\textbf{Initialization}: We use Kaiming He initialization \citep{he2015delving} for all weight matrices:
\begin{equation}
W_{ij} \sim \mathcal{N}\left(0, \frac{2}{n_{\text{in}}}\right)
\end{equation}
where $n_{\text{in}}$ is the number of input units to the layer.

\subsubsection{Training Hyperparameters}

\begin{table}[h]
\centering
\begin{tabular}{ll}
\toprule
\textbf{Parameter} & \textbf{Value} \\
\midrule
Optimizer & AdamW \\
$\beta_1, \beta_2$ & 0.9, 0.98\\
$\epsilon$ & $10^{-8}$ \\
Learning rate & $1 \times 10^{-3}$\\
Batch size & 256 \\
Epochs & 200 \\
Weight decay & $10^{-4}$ \\
Loss function & Cross-entropy \\
\bottomrule
\end{tabular}
\caption{Complete hyperparameters for MNIST MLP training.}
\label{tab:mnist_hyperparams}
\end{table}

\textbf{Data normalization}: MNIST images are normalized using dataset statistics: mean = 0.1307, std = 0.3081.

\textbf{Hardware}: Single NVIDIA H100 GPU (80GB).

\textbf{Implementation}: PyTorch 2.0 with PyTorch Lightning for training loops.

\textbf{Random seeds}: For each configuration ($h \in \{64, 128, 256\}$), we train 5 independent models with random seeds 0--4.

\subsection{Jacobian Computation Details}

\subsubsection{Vectorized Jacobian Calculation}

For a batch of inputs $X \in \mathbb{R}^{B \times d}$ and hidden representation function $\phi: \mathbb{R}^d \to \mathbb{R}^h$, we compute the Jacobian for each sample using PyTorch's automatic differentiation:

\begin{equation}
J_i = \frac{\partial \phi(x_i)}{\partial x_i} \in \mathbb{R}^{h \times d}, \quad i = 1, \ldots, B
\end{equation}

We use \texttt{torch.autograd.functional.jacobian} with vectorization enabled for efficiency. The resulting tensor has shape $(B, h, d)$, containing $B$ Jacobian matrices.

\subsubsection{Memory-Efficient Minor Computation}

We compute the \textbf{full Jacobian matrix} $J \in \mathbb{R}^{h \times d}$ for each layer using vectorized automatic differentiation.

However, computing all possible $k$-minors from the Jacobian is prohibitively expensive, as it requires selecting all $\binom{h}{k}$ row combinations and all $\binom{d}{k}$ column combinations. For example, MNIST L0 ($h=256$, $d=784$) has $\binom{256}{3} \times \binom{784}{3} \approx 2.2 \times 10^{15}$ possible 3-minors.

We therefore employ random sampling of row and column combinations:

\textbf{Row and column sampling}: For $k > 1$, we randomly sample up to 10,000 row combinations (from $\binom{h}{k}$ possibilities) and up to 10,000 column combinations (from $\binom{d}{k}$ possibilities). Sampling is deterministic (seed=42) for reproducibility.

\textbf{Chunked processing}: We split the sampled combinations into chunks of 500 and process them sequentially, clearing GPU memory between chunks to avoid out-of-memory (OOM) errors.

\subsubsection{Layer-wise Jacobian Analysis}

For a 2-layer MLP, we compute Jacobians at each hidden layer independently:
\begin{itemize}
    \item \textbf{Layer 1 (L1)}: $J_1 = \frac{\partial \phi_1}{\partial x}$ where $\phi_1(x) = \sigma(W_0 x + b_0) \in \mathbb{R}^h$
    \item \textbf{Layer 2 (L2)}: $J_2 = \frac{\partial \phi_2}{\partial x}$ where $\phi_2(x) = \sigma(W_1 \phi_1(x) + b_1) \in \mathbb{R}^h$
\end{itemize}

\subsection{Evaluation Protocol Details}

\subsubsection{Timeline Analysis}

We compute all KA geometry metrics every 10 epochs. Epoch 0 corresponds to random initialization and serves as the baseline.

\subsubsection{Random Initialization Baseline}

For each trained model, we save a copy of the randomly initialized network (before any training) and compute the same Jacobian metrics. This allows us to measure the \emph{change} in KA signatures due to training rather than properties of the initialization scheme.

\subsubsection{Statistical Robustness}

We report mean $\pm$ standard deviation across 5 random seeds for each configuration ($h \in \{64, 128, 256\}$).

\subsubsection{Sampling Parameters for Metrics}

To balance statistical accuracy with computational cost, we use the following sampling strategies:

\begin{table}[h]
\centering
\begin{tabular}{ll}
\toprule
\textbf{Parameter} & \textbf{Value} \\
\midrule
Evaluation batches & 2 (512 samples) \\
Rotation samples & 400 \\
Row samples (for minors) & 10,000 \\
Column samples (for minors) & 10,000 \\
Max minor order $k$ & 3 \\
\bottomrule
\end{tabular}
\caption{Sampling parameters for computing KA geometry metrics. Note: We compute the \emph{full} Jacobian matrix but sample row and column combinations when computing minors (determinants).}
\label{tab:sampling_params}
\end{table}

\subsection{Computational Resources}

Each seed run takes approximately 2--3 hours on a single H100 GPU (including training and Jacobian analysis).

\subsection{Software Versions}

\begin{table}[h]
\centering
\begin{tabular}{ll}
\toprule
\textbf{Package} & \textbf{Version} \\
\midrule
Python & 3.10 \\
PyTorch & 2.0.1 \\
PyTorch Lightning & 2.0.0 \\
torchvision & 0.15.2 \\
CUDA & 12.1 \\
\bottomrule
\end{tabular}
\caption{Software versions used for all experiments.}
\label{tab:software_versions}
\end{table}